\documentclass[10pt,twocolumn,letterpaper]{article}
\usepackage[pagenumbers]{cvpr}

\usepackage{amsmath}
\usepackage{amssymb}
\usepackage{booktabs}
\usepackage{pifont}
\usepackage{marvosym}
\usepackage{multirow}
\usepackage{xspace}
\usepackage{xcolor}
\usepackage{tabularx}
\usepackage{graphicx}
\usepackage{subcaption}
\usepackage{url}
\newcommand{\name}{WorldGrow\@\xspace}
\def\eg{\emph{e.g}\onedot} 
\def\ie{\emph{i.e}\onedot}

\makeatother
\definecolor{metric-1}{HTML}{ff9999}
\definecolor{metric-2}{HTML}{ffcc99}
\definecolor{metric-3}{HTML}{fff6b2}

\definecolor{cvprblue}{rgb}{0.21,0.49,0.74}
\usepackage[pagebackref,breaklinks,colorlinks,allcolors=cvprblue]{hyperref}

\title{\name: Generating Infinite 3D World}

\author{
Sikuang Li\textsuperscript{\rm 1}\footnotemark[1] \footnotemark[2] \quad 
Chen Yang\textsuperscript{\rm 2}\footnotemark[1]  \quad
Jiemin Fang\textsuperscript{\rm 2}$^\text{\Letter}$ \quad 
Taoran Yi\textsuperscript{\rm 3}\footnotemark[2] \quad 
Jia Lu\textsuperscript{\rm 3}\footnotemark[2]\\
\vspace{5pt}
Jiazhong Cen\textsuperscript{\rm 1}\footnotemark[2] \quad 
Lingxi Xie\textsuperscript{\rm 2} \quad 
Wei Shen\textsuperscript{\rm 1} \quad 
Qi Tian\textsuperscript{\rm 2}$^\text{\Letter}$\\
\textsuperscript{\rm 1}MoE Key Lab of Artificial Intelligence, School of Computer Science, SJTU\\
\vspace{5pt}
\textsuperscript{\rm 2}Huawei Inc. \quad
\textsuperscript{\rm 3}Huazhong University of Science and Technology\\
\url{World-Grow.github.io}
}

\begin{document}
% \maketitle
\twocolumn[{%
\renewcommand\twocolumn[1][]{#1}%
\maketitle
\vspace{-2em}
\begin{center}
\centering
\includegraphics[width=1.0\linewidth]{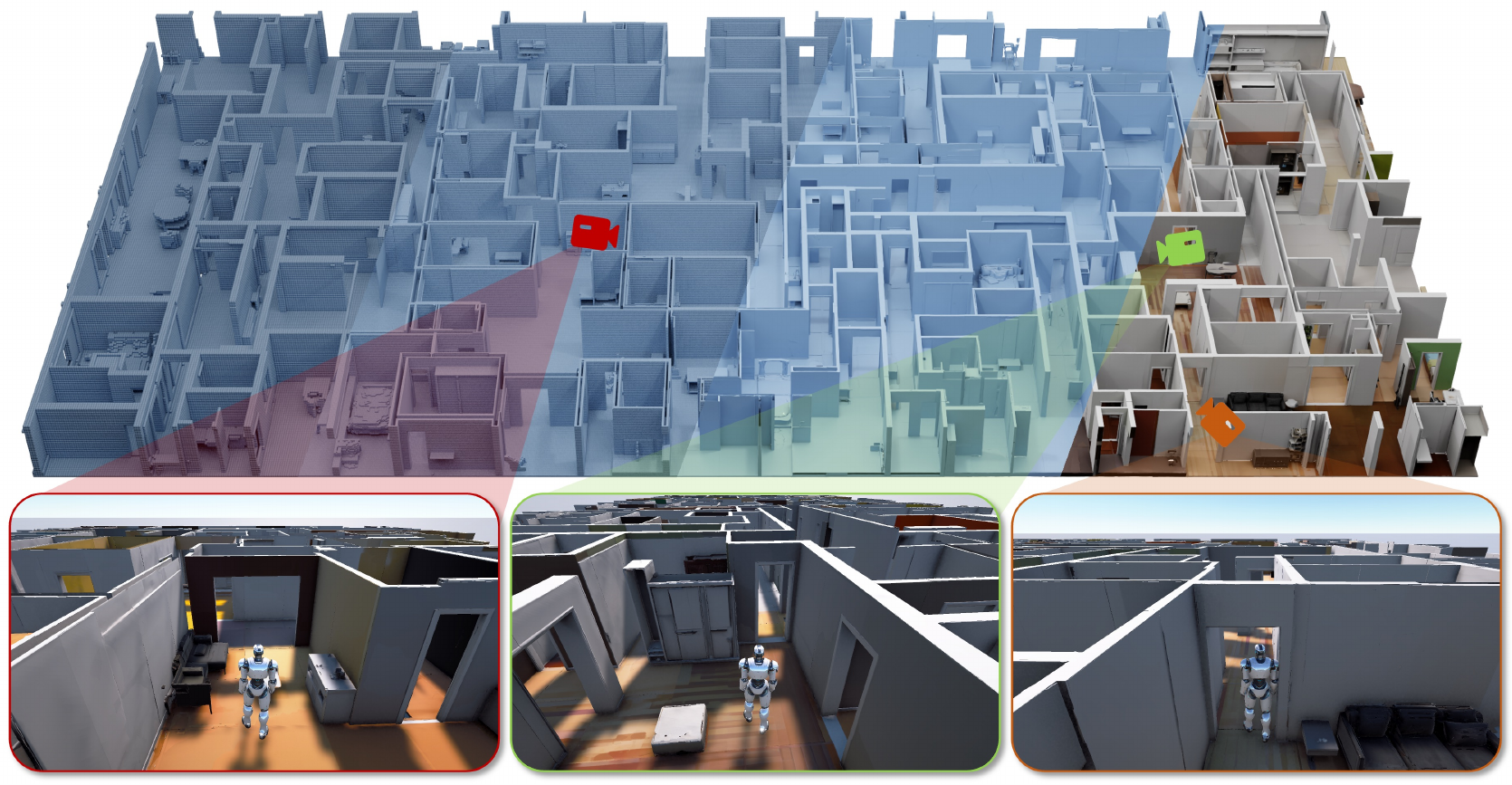}
\vspace{-2em}
\captionsetup{hypcap=false}
\captionof{figure}{We introduce \textbf{WorldGrow}, a novel framework for infinite 3D world generation via block-wise synthesis and growth with coarse-to-fine refinement. Starting from a single seed block, \name progressively generates large-scale 3D scenes with coherent geometry and photorealistic appearance. \emph{Top:} An indoor scene of \(19\times39\) blocks (covering \(\sim\)1{,}800\,m\(^2\)); left→right: coarse voxel layout, refined voxels, mesh reconstruction, and textured rendering. \emph{Bottom:} An embodied agent navigates the generated world, demonstrating diverse room layouts and traversable spaces suitable for interactive AI tasks (\eg, navigation and planning).
}
\label{fig:teaser}
\end{center}
}]

{
\renewcommand{\thefootnote}{\fnsymbol{footnote}}
\footnotetext[1]{Equal contributions.}
\footnotetext[2]{Work done during internship at Huawei.}
\footnotetext[0]{\hspace{-5pt}$^\text{\Letter}$Corresponding authors.}
}

\begin{abstract}
We tackle the challenge of generating the infinitely extendable 3D world -- large, continuous environments with coherent geometry and realistic appearance. Existing methods face key challenges: 2D-lifting approaches suffer from geometric and appearance inconsistencies across views, 3D implicit representations are hard to scale up, and current 3D foundation models are mostly object-centric, limiting their applicability to scene-level generation. 
Our key insight is leveraging strong generation priors from pre-trained 3D models for structured scene block generation. To this end, we propose \textbf{\textit{\name}}, a hierarchical framework for unbounded 3D scene synthesis.
Our method features three core components: (1) a data curation pipeline that extracts high-quality scene blocks for training, making the 3D structured latent representations suitable for scene generation; (2) a 3D block inpainting mechanism that enables context-aware scene extension; and (3) a coarse-to-fine generation strategy that ensures both global layout plausibility and local geometric/textural fidelity.
Evaluated on the large-scale 3D-FRONT dataset, \name achieves SOTA performance in geometry reconstruction, while uniquely supporting infinite scene generation with photorealistic and structurally consistent outputs. These results highlight its capability for constructing large-scale virtual environments and potential for building future world models.
\end{abstract}

\section{Introduction}
This paper addresses the critical challenge of generating the infinitely extendable 3D world, aiming to automatically create vast, continuous, and content-rich virtual environments. Such technology holds significant potential for industries including video games, virtual/augmented reality (VR/AR), computer-aided design, and film production. More importantly, infinite 3D world generation is foundational for developing \textit{World Models} and embodied AI systems~\cite{citydreamer,uniscene}, as it provides continuously expandable environments essential for open-ended learning, where agents can navigate, plan, and interact without the constraints of fixed-size worlds.

To achieve infinite 3D world generation, existing efforts have primarily explored two main approaches. One line of works~\cite{wonderjourney,luciddreamer,syncity,worldlabs} relies on pre-trained 2D diffusion models~\cite{stablediffusion,flux,genie3} to generate images, which are then ``lifted'' to 3D scenes using camera poses, depth maps~\cite{midas,midas3_1}, or image-to-3D models~\cite{trellis}.
These methods optimize based on local viewpoints and lack a holistic understanding of the full 3D structure. As a result, they often suffer from geometric inaccuracies and appearance inconsistencies (\eg, aliasing or distortion) across different views or extended regions, which further limits their ability to generate large-scale scenes.
Another line of works~\cite{blockfusion,lt3sd,nuiscene} attempts to directly predict 3D representations (\eg, triplanes~\cite{eg3d,blockfusion}, UDFs~\cite{udf,tudfnerf,lt3sd}, global latents~\cite{diffusionsdf}) by learning from 3D data for scene generation. However, their performance and generalization are often constrained by the limited scale and diversity of available scene-level datasets~\cite{3dfront,urbanscene3d}. 
Recent powerful 3D generation models~\cite{trellis,clay,craftsman,lion}, empowered by large-scale training data~\cite{objaversexl}, have demonstrated impressive capabilities in producing high-quality 3D assets. Though powerful, they are predominantly designed for single object generation, not applicable for infinite scene generation.

We propose to leverage the powerful generative capabilities of 3D generation models for block-based infinite scene generation -- a promising yet challenging direction. The key challenges are threefold: 1) transferring rich geometric and textural priors from object-level models to generate scene blocks that are contextually coherent, rather than isolated assets;
2) ensuring seamless geometric, stylistic, and textural coherence between adjacent 3D blocks during iterative scene growth;
3) achieving global structural plausibility and semantic diversity in large-scale compositions, avoiding incoherent arrangements.

To address these challenges, we introduce \name, a novel framework that, for the first time, enables the generation of infinite continuous 3D \underline{World}s with plausible layouts and high-fidelity appearances in a region-\underline{grow}ing manner.
First, we design a data preparation pipeline to extract sufficient high-quality ground-truth scene blocks from existing datasets. In addition, we adapt object-level 3D representation to be scene-friendly, enabling the migration of learned object priors for generating scene blocks with fine-grained geometry and appearance.
Second, we develop a 3D block inpainting pipeline to ensure robust and context-aware completion of missing blocks during iterative extension.
Finally, to ensure both global coherence and local detail, we curate coarse and fine datasets focused on layouts and appearances, respectively. During generation, a coarse-trained model builds the scene structure first, then a fine-trained model refines detailed geometry and textures.
As shown in Fig.~\ref{fig:teaser}, \name generates detail-rich, photorealistic, and infinitely extendable 3D scenes, highlighting its strong potential for large-scale virtual world construction.

In summary, our main contributions are as follows:

1) A systematic data construction pipeline and the created scene block datasets, enabling scalable training and evaluation for block-based infinite scene generation.

2) An infinite 3D scene generation framework, \name, which synthesizes continuous and unbounded 3D worlds with coherent layouts and photorealistic appearances.

3) A set of novel techniques enabling high-quality world generation, including scene-friendly SLATs for adapting object-level priors, a 3D inpainting method for seamless block completion, and a coarse-to-fine generation strategy that balances the global structure and local details.

\section{Related Work}

\subsection{3D Generation Pretrained Models}

Recent advances in 3D pretraining have shown great promise in single-image 3D object generation. Leveraging representations such as triplanes~\cite{eg3d} and 3D Gaussian Splatting (3DGS)\cite{3dgs}, a number of feed-forward models\cite{lrm,pflrm,meshlrm,ln3diff,3dtopia,triplanegaussian,dreamgaussian} have been developed to directly synthesize 3D content from a single image.

\begin{figure*}[thbp]
\centering
\includegraphics[width=\linewidth]{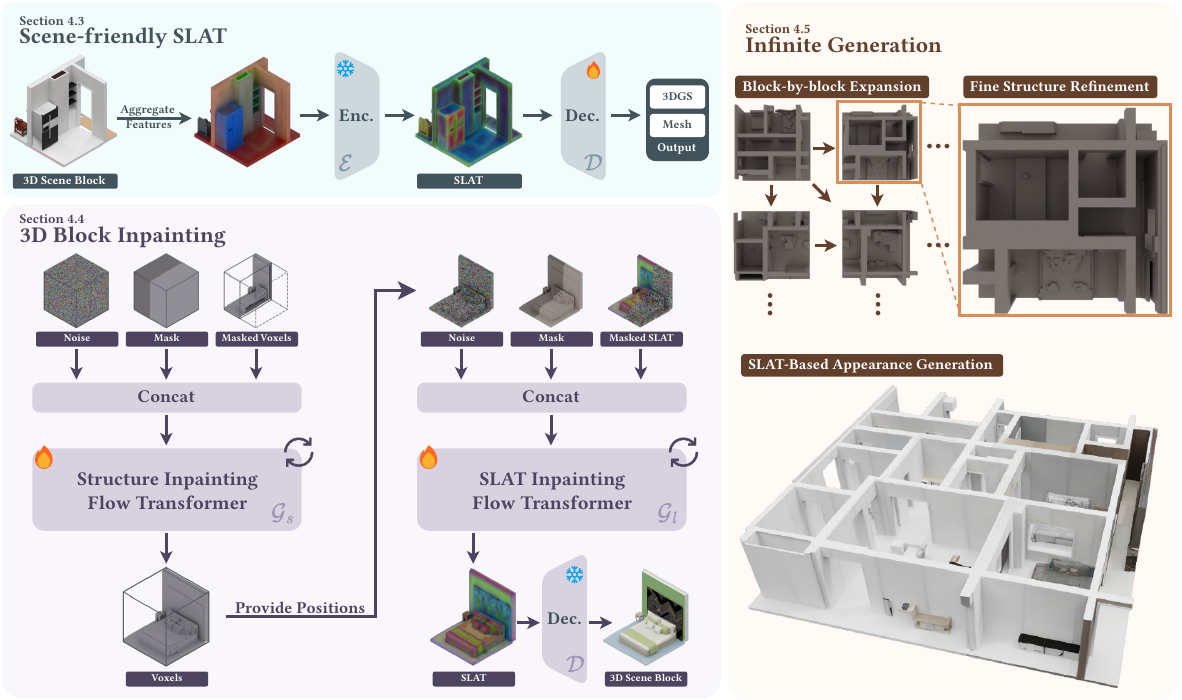}
\caption{
Overview of \name. Our goal is to generate infinite 3D scenes through modular, block-by-block synthesis. We begin by curating high-quality scene blocks and adapting SLAT to better model structured 3D context. A 3D block inpainting module enables spatially coherent extension, while a coarse-to-fine generation strategy ensures global layout plausibility and local detail fidelity. Together, these components allow \name to progressively construct photorealistic and structurally consistent 3D worlds with theoretically unbounded spatial extent.
}
\label{fig:pipe}
\end{figure*}

To better capture 3D structural priors, several works~\cite{3dtopiaxl,clay,craftsman,direct3d,xcube,lion,michelangelo,sdfdiffusion} focus on sampling voxels or point clouds from 3D shapes to coarsely define geometry, which are then embedded into latent spaces~\cite{vae} via generative models~\cite{ddpm,ddim}. TRELLIS~\cite{trellis} introduces a novel 3D representation named Structured LATents (SLAT), which encodes object shapes into sparse voxel grids with DINOv2 features~\cite{dinov2}. Later extensions~\cite{triposf,hunyuan3d,direct3ds2,ultra3d} improve SLAT by revising voxel sampling strategies. UniLat3D~\cite{unilat3d} unifies geometry and appearance into a single compact latent space for direct single-stage generation.
These methods are typically trained on large-scale 3D object datasets~\cite{objaversexl,abo,3dfuture,hssd,toys4k}, achieving high-quality object reconstruction. However, their applicability to scene-level generation is still underexplored, and we perform some preliminary attempts in this paper.

\vspace{10pt}
\subsection{3D Unbounded Scene Generation}

To extend 3D generation to scene-level tasks, recent works explore the use of 2D inpainting diffusion models~\cite{wonderjourney,luciddreamer} or video diffusion models with camera control~\cite{svd,dynamicrafter,motionctrl,viewcrafter,wonderland} to hallucinate plausible multi-view images from single-view or textual inputs. Other approaches~\cite{text2room,diffindscene,dreamscene,hiscene,diffuscene} aim to generate more complete geometry, such as 360-degree scenes within indoor environments. However, these methods typically generate only small-scale scenes, such as a single room.

More recent efforts toward unbounded scene generation aim to produce 3D content that can be extended infinitely in all directions. BlockFusion~\cite{blockfusion} partitions 3D scenes into local blocks, encodes them as triplanes, and employs triplane extrapolation to synthesize neighboring blocks. Other methods utilize Truncated Unsigned Distance Field (TUDF)~\cite{udf,tudfnerf,lt3sd} or vector-set latents~\cite{nuiscene} to reconstruct the 3D scene blocks.

While these methods achieve compelling unbounded geometry generation, they typically lack explicit texture modeling. Instead, they rely on external texture synthesis and mapping pipelines~\cite{scenetex,dreamspace} to produce realistic surface appearances. SynCity~\cite{syncity} proposes a training-free pipeline that divides a scene into grids, generates descriptive captions for each grid using Large Language Models (LLMs)~\cite{gpt4}, synthesizes images via text-to-image diffusion models~\cite{flux}, and finally reconstructs textured 3D scenes using pretrained 3D generation models. Despite its scalability, this method suffers from limited view consistency: high-quality rendering is restricted to camera poses seen during diffusion generation, with fidelity degrading as the viewpoint diverges.

\section{Preliminary: TRELLIS} \label{sec:prelim}
TRELLIS~\cite{trellis}, serving as a foundational model to our work, is a text/image-conditioned 3D generation model operating by denoising features in a sparse 3D latent space.

\paragraph{Structured Latent Representation.}
TRELLIS represents 3D objects via structured latents (SLATs):
$
\mathbf{z} = \{(\mathbf{z}_i, \mathbf{p}_i)\}_{i=1}^{L}
$.
Here, $\mathbf{z}_i \in \mathbb{R}^C$ is a latent feature at position $\mathbf{p}_i \in \{0, \dots, N-1\}^3$ ($N$: grid resolution), and $L \ll N^3$ is the count of active surface voxels. SLATs encode coarse geometry structure and fine appearance of 3D objects by linking latent features to active voxel locations using a Transformer-based variational autoencoder (VAE)~\cite{vae}, including an encoder $\mathcal{E}$ and a decoder $\mathcal{D}$.
$\mathcal{E}$ maps sparse voxel features $\mathbf{f} = \{(\mathbf{f}_i, \mathbf{p}_i)\}_{i=1}^L$ of a 3D object to structured latents $\mathbf{z}$, where $\mathbf{f}_i$ is a local visual feature created by projecting DINOv2~\cite{dinov2} feature maps (derived from multiview renders of the object) on to the voxel $\mathbf{p}_i$ and averaging the retrieved features. $\mathcal{D}$ then decodes $\mathbf{z}$ into 3D representations, \eg, 3D Gaussians~\cite{3dgs}, radiance fields~\cite{nerf}, and meshes.

\paragraph{Structured Latent Generation.}
SLAT generation is a two-stage pipeline: Stage~1 predicts active voxel centers $\{\mathbf{p}_i\}_{i=1}^L$, and stage~2 recovers their latent features $\{\mathbf{z}_i\}_{i=1}^L$. 
Each stage uses a flow Transformer~\cite{flow2} $\upsilon_\theta$ on a latent code $\mathbf{\ell}$. 
It learns to reverse noise addition ($\mathbf{\ell}^{(t)}=(1-t)\mathbf{\ell}^{(0)}+t\mathbf{\epsilon}$, where $\epsilon\sim\mathcal{N}(0,\mathbf{I})$ and $t\in[0,1]$) by minimizing the flow loss:
\[
\label{eq:diff}
\min_\theta\mathbb{E}_{(\mathbf{\ell}^{(0)},x), t,\mathbf{\epsilon}}
\|
\upsilon_\theta(\mathbf{\ell}^{(t)}, x,t)
- (\mathbf{\epsilon}-\mathbf{\ell}^{(0)})
\|^2_2,
\]
where $x$ is a conditioning prompt which can be either an image or a text prompt.
In Stage~1 (for $\{\mathbf{p}_i\}$), $\mathbf{\ell}\in\mathbb{R}^{L'\times C'}$ comprises $L'$ tokens from a compressed $N^3$ occupancy volume.
In Stage~2 (for $\{\mathbf{z}_i\}$), $\mathbf{\ell} = \{\mathbf{z}_i\}_{i=1}^L \in \mathbb{R}^{L\times C}$ is a matrix of $L$ $C$-dimensional tokens ($L$: active voxels from Stage~1). For better understanding, we term the flow Transformer in stage~1 as structure generation $\mathcal{G}_s$ and flow Transformer in stage~2 as latent generation $\mathcal{G}_l$.

\section{Method}
\label{sec:method}

\subsection{Task Definition and Overall Framework}

We define the task of synthesizing an infinite 3D world $\mathcal{W}$ exhibiting plausible layouts and high-fidelity appearances as follows.
The world $\mathcal{W}$ is conceptualized as an unbounded composition of interconnected 3D blocks. Each block $\mathcal{B}$ within this world is generated iteratively, conditioned on previously synthesized blocks.
For simplicity, we define $\mathcal{B}$ as a rectangular block aligned with the horizontal axes (\ie, $XY$), with equal widths in the $X$ and $Y$ directions.

To enable infinite scene generation, \name first curates high-quality scene blocks for training (Sec.~\ref{sec:data}). We adapt the SLAT representation for structured 3D block modeling (Sec.~\ref{sec:scene}), implement a 3D block inpainting module for context-aware completion (Sec.~\ref{sec:inpainting}), and describe our coarse-to-fine generation strategy that achieves global layout plausibility with local detail fidelity (Sec.~\ref{sec:inf}). The pipeline is shown in Fig.~\ref{fig:pipe}.

\subsection{Data Curation}\label{sec:data}
To enable infinite scene generation, we begin by constructing a dataset of structured, extendable 3D blocks. Existing 3D datasets, such as Objaverse-XL~\cite{objaversexl}, are predominantly object-centric, consisting of isolated assets without spatial continuity. TRELLIS~\cite{trellis} performs well on such object-level data, but is not applicable to scene-level generation, which requires modular units that are spatially aligned and context-aware.

\paragraph{Scene Slicing.} \label{sec:slice}
To bridge this gap, we propose a scene slicing strategy that partitions full 3D scenes (\eg, a house or city) into coherent and reusable blocks. 
Given a full scene mesh, we extract training-ready blocks through the following process: we import the mesh into Blender~\cite{blender}, place a cuboid within its bounding box, and extract content via Boolean Intersection with the scene geometry. 
To ensure spatial density and avoid sparse regions, we render a top-down view and compute the occupancy of each extracted cuboid—if less than 95\% of the surface contains visible content, the cuboid is repositioned and re-evaluated. 
This iterative sampling process yields multiple valid placements per scene, constructing a diverse set of spatially dense scene blocks. Our curated, topologically consistent data significantly reduces unrealistic geometry compared to naive partitioning approaches.

\paragraph{Coarse-to-Fine Data Strategy.}
Our method aims to synthesize unbounded, high-fidelity virtual worlds composed of 3D scene blocks with plausible global layouts. However, each block must be encoded into a SLAT, whose limited representational capacity constrains the amount of geometry and appearance detail it can effectively preserve. This introduces a fundamental trade-off in block design: larger 3D blocks capture broader scene context, benefiting global layout learning, but may suffer in rendering fidelity; conversely, smaller blocks support finer visual quality but lack sufficient spatial context to learn coherent scene structures.

To address this, we adopt a coarse-to-fine data strategy that balances context and detail. We prepare two distinct datasets: \textit{coarse} and \textit{fine} blocks\footnote{Throughout this paper, superscripts $c$ and $f$ on symbols denote their association (typically via training or definition) with the coarse and fine datasets, respectively.}. Coarse blocks are defined with four times the area in the $XY$ plane while maintaining the same height, thereby capturing larger spatial volumes and richer contextual information. Both types of blocks are extracted using the random spatial partitioning method described previously. These dual-resolution datasets form the foundation for training our generative pipeline across global layout generation and local detail refinement.

\subsection{Scene-friendly SLAT} \label{sec:scene}
While SLAT has demonstrated strong performance in object-level generation, its direct application to 3D scene block synthesis faces critical limitations. We identify two primary challenges:
1) Direct feature aggregation. SLAT's VAE training projects multiview DINOv2 features onto each voxel $\mathbf{p}_i$ and aggregates them to form its visual feature $\mathbf{f}_i$. While effective for objects with minimal self-occlusion, this projection-based aggregation degrades in cluttered scenes where self-occlusions are prevalent. As a result, vanilla SLAT often fails to capture accurate spatial relationships, leading to artifacts such as color bleeding between adjacent surfaces.
2) Inadequate decoder for scene blocks. SLAT's decoder $\mathcal{D}$ is pre-trained on object-level data, which typically lacks detailed 3D content near object boundaries. As a result, when applied to scene blocks, $\mathcal{D}$ often produces floaters and artifacts near the block edges. These decoding failures lead to visual discontinuities, such as floating geometry or broken transitions, when multiple blocks are composed into large-scale scenes.

\begin{figure}[tbp]
\centering
\includegraphics[width=\columnwidth]{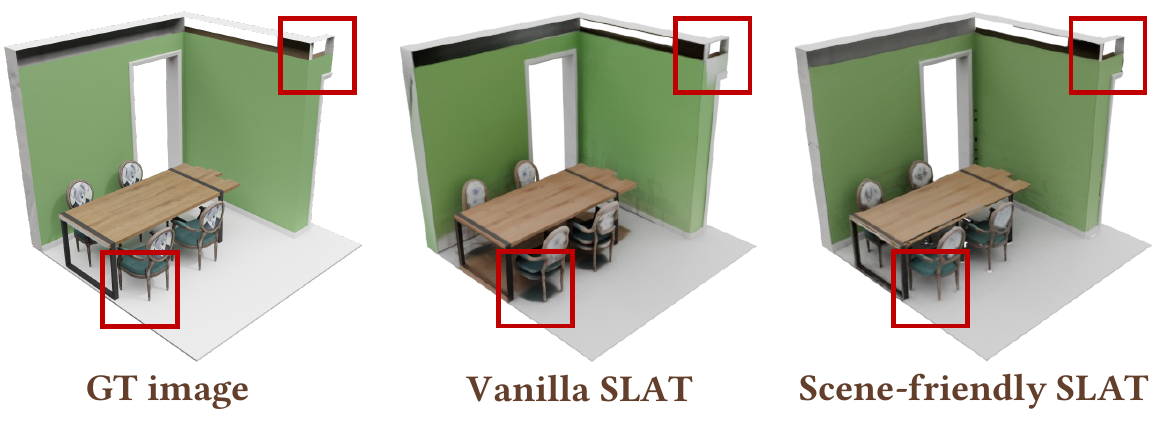}
\caption{Scene-friendly SLAT better models 3D scene blocks, particularly in areas with occlusions and near block boundaries.}
\label{fig:scene-friendlySLAT}
\end{figure}

To address these limitations, we introduce two key modifications to make SLAT more scene-friendly. First, we incorporate an occlusion-aware strategy during feature aggregation. While conceptually simple, this adjustment significantly improves the representation of occluded regions and yields more consistent voxel features in cluttered scenes. 
Second, we retrain the decoder $\mathcal{D}$ on scene block data, shifting its focus from isolated objects to structured scene content. This adaptation enables the decoder to better handle boundary regions, resulting in cleaner geometry and more coherent textures, especially at block edges.
Together, these adaptations substantially reduce structural artifacts and enable more reliable scene block synthesis, as shown in Fig.~\ref{fig:scene-friendlySLAT}.

\subsection{3D Block Inpainting}\label{sec:inpainting}
While scene-friendly SLAT improves the quality and consistency of individual block synthesis, extending a scene block-by-block requires reasoning over partial context and ensuring continuity with surrounding geometry and appearance. To address this challenge, we formulate scene expansion as a 3D block inpainting task, where a missing target block is synthesized based on its surrounding spatial neighbors.

Inherent from TRELLIS, we use a two-stage inpainting framework that operates on structure and latent space. 
Given a partially observed block with missing regions, our model first predicts the 3D structure (\(\mathbf{p}_b\)) and then reconstructs the corresponding latent features (\(\mathbf{z}_b\)) for high-fidelity appearance synthesis. To enable the model to better localize and infer missing regions, we modify the input layer of the models. Specifically, instead of using noisy latents as input, we concatenate three components along the channel dimension: the noisy latents, a binary mask indicating the inpainting region, and the masked known region itself. This design allows the model to condition its prediction on both the known context and explicit spatial cues of the missing area. By learning to denoise this composite input, the network is able to infer the structure and appearance of missing regions while preserving the observed content, improving the spatial continuity and stability of 3D block inpainting.

To train the inpainting model, we randomly select two splitting positions along the \(X\) and \(Y\) axes to divide each scene block into four quadrants, keeping one as context and masking the remaining three. For \textit{structure inpainting}, we define a voxel-level binary mask \(m_s \in \{0,1\}^{N \times N \times N}\), where \(m_s=1\) denotes voxels to be inpainted. The structure generator \(\mathcal{G}_s\) takes this mask as input to complete the missing geometry. For \textit{latent inpainting}, we define a sparse mask \(m_l = \{(m_i, \textbf{p}_i)\}_{i=1}^{L}\), where \(\textbf{p}_i\) is the spatial coordinate and \(m_i \in \{0,1\}\) indicates whether to inpaint. This guides the latent generator \(\mathcal{G}_l\) to reconstruct corresponding features.

\begin{figure}[thbp]
\centering
\includegraphics[width=\columnwidth]{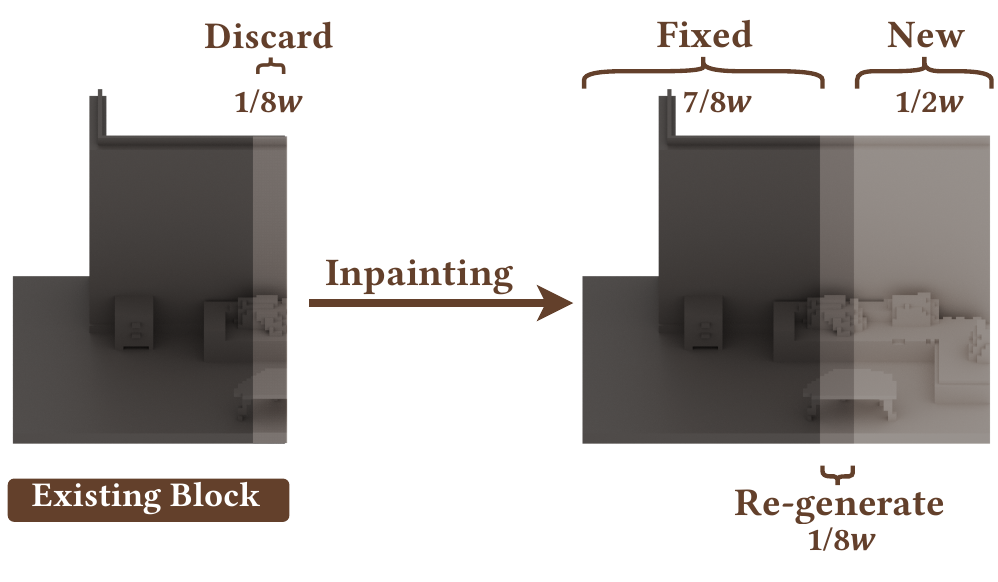}
\vspace{-2em}
\caption{
1D illustration of our block-by-block expansion. Existing block's $[1/2w,\; 7/8w)$ area is used as context for inpainting the next block. Thus, the final region $[7/8w,\; w)$ is discarded and then re-generated during expansion.
}
\label{fig:blockbyblock}
\end{figure}

Both generators are optimized using a flow-matching loss:
\[
\label{eq:inpainting_loss}
\min_{\theta}\,
\mathbb{E}_{(\boldsymbol{\ell}^{(0)},m,x),\,t,\,\boldsymbol{\epsilon}}
\|
\mathcal{G}(\boldsymbol{\ell}^{(t)}, m, \boldsymbol{\ell}^{(0)}_m,\; x,\; t)
- (\boldsymbol{\epsilon} - \boldsymbol{\ell}^{(0)})
\|_{2}^{2},
\]
where \(\boldsymbol{\ell}^{(0)}_m=\boldsymbol{\ell}^{(0)} \otimes (1 - m)\) is the latent code masked, \(\otimes\) denotes the Hadamard product, and \((\mathcal{G}, \boldsymbol{\ell}, m)\) corresponds to either \((\mathcal{G}_s, \mathbf{p}_b, m_s)\) or \((\mathcal{G}_l, \mathbf{z}_b, m_l)\), depending on the task.

To support coarse-to-fine generation, we train separate models: $\mathcal{G}_s^c$ on coarse blocks for structure inpainting, and $\mathcal{G}_s^f$, $\mathcal{G}_l^f$ on fine blocks for structure and latent inpainting, respectively -- balancing global coherence and local detail.

\begin{figure*}[thbp]
\centering
\includegraphics[width=\linewidth]{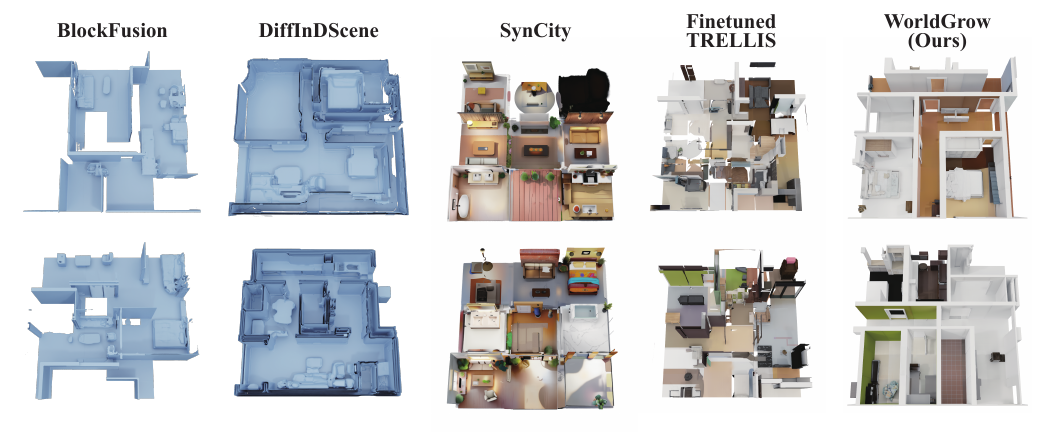}
\vspace{-2em}
\caption{Qualitative comparison of indoor scene generation. We compare our method with state-of-the-art infinite scene generation approaches, indoor house generation methods and our baseline TRELLIS. \name produces high-resolution, continuous indoor scenes with realistic and coherent textures.}
\label{fig:qualitative}
\end{figure*}

\subsection{Infinite Scene Generation} \label{sec:inf}
With all components available, we now describe how \name constructs an infinite 3D world $\mathcal{W}$ via a block-based, coarse-to-fine generation strategy. Starting from a seed block, the world is progressively extended in the $XY$ plane through iterative 3D block inpainting. 
A coarse model first lays out the global structure across blocks, which is then refined by fine-level models to recover detailed geometry and generate the corresponding SLATs for each region. 

\paragraph{Block-by-Block Expansion.}
We initiate scene generation from a seed block, which can either be synthesized by our inpainting model with a full 3D mask or initialized using a sample from vanilla TRELLIS. The scene is then expanded block by block, typically along the $+X$ and $+Y$ directions.

For each new block, the inpainting model takes as context the previously generated blocks to its left, top, and top-left (if available). See Fig.~\ref{fig:blockbyblock} for a 1D illustration of the expansion. To ensure continuity, we reuse a portion of these existing blocks: Specifically, we reuse a $3/8w$-wide margin from each neighboring block along $X$ and $Y$ axes. This overlapping region corresponds to $[1/2w,\; 7/8w)$ on each axis. Based on this context, we inpaint the central $5/8w \times 5/8w$ region to complete a new $12/8w \times 12/8w$ block. This overlapping design ensures smooth transitions across block boundaries and provides a consistent context window for each expansion step.

\paragraph{Coarse Structure Generation.}
To establish the large-scale layout of the scene, we first apply the block-by-block generation process using the coarse structure model $\mathcal{G}_s^c$. This produces a low-resolution but spatially coherent structure $\mathbf{p}_w^c$ that defines the overall geometry of the world.

\paragraph{Fine Structure Refinement.}
To enrich local geometry, we refine $\mathbf{p}_w^c$ using the fine structure generator $\mathcal{G}_s^f$. We begin by upsampling $\mathbf{p}_w^c$ via trilinear interpolation to match the voxel resolution of the fine stage, producing $\mathbf{p}_w^{c \uparrow f}$. This high-resolution structure is then partitioned into standard fine blocks.

Rather than generating each fine block from scratch, we adopt a structure-guided denoising approach inspired by SDEdit~\cite{sdedit}. For each upsampled fine block \(\mathbf{p}_{\text{fblock}}^{c \uparrow f}\), we encode it into an initial latent \(\mathbf{\ell}^{(0)}_{\text{fblock}}\). We then perturb this latent with controlled Gaussian noise:
\[
\mathbf{\ell}^{(t')}_{\text{fblock}} = (1 - t') \mathbf{\ell}^{(0)}_{\text{fblock}} + t' \boldsymbol{\epsilon}, \quad \text{where } 0 < t' < t.
\]
The fine generator \(\mathcal{G}_s^f\) denoises \(\mathbf{\ell}^{(t')}_{\text{fblock}}\) to reconstruct the refined structure \(\mathbf{p}^f_{\text{fblock}}\). This strategy enables preserving space distribution priors while enhancing details, effectively bridging global layout and fine-scale realism in a structure-aware generation process.

\paragraph{SLAT-Based Appearance Generation.}
Once the fine-level structure of the world $\mathcal{W}$, denoted as $\mathbf{p}_w^f$, is complete, we generate the corresponding SLATs $\mathbf{z}_w$. This stage follows the same block-by-block generation strategy as used for structure, but operates in the latent space. For each block, the latent generator $\mathcal{G}_l^f$ synthesizes latents based on previously generated SLAT and current structure mask. Unlike structure inpainting, which uses dense voxel masks, latent inpainting is guided by sparse latent masks. After all latent blocks are generated, the full SLAT $\mathbf{z}_w$ is decoded by our retrained $\mathcal{D}$ into a renderable 3D world $\mathcal{W}$. 

\section{Experiments}
\subsection{Experiment Settings} \label{sec:impl}
\begin{figure}[t]
\centering
\includegraphics[width=\columnwidth]{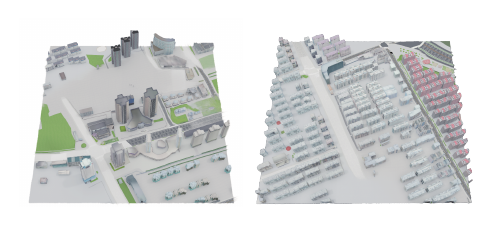}
\caption{Infinite outdoor 3D scene generation by \name. Our method synthesizes diverse scenes such as urban streetscapes with plausible layouts, coherent suburban neighborhoods with consistent styles, showing \name's ability to be adapted to various domains.}
\label{fig:outdoor}
\end{figure}

\paragraph{Datasets.}
To align with previous infinite generation methods, We train \name on the dataset processed from 3D-FRONT~\cite{3dfront,3dfuture}. From the original 6,811 houses, we retain 3,072 after filtering, and include 353 additional houses that were manually corrected for higher quality.
Consequently, our final dataset comprises 3,425 curated houses with reasonable layouts and detailed furnishings.
From these, we generate 120k fine blocks and 38k coarse blocks.
We also verify \name with city dataset UrbanScene3D~\cite{urbanscene3d} in Fig.~\ref{fig:outdoor}. Please refer to the Appendix for details.

\begin{table}[t]
\centering
\small
\setlength{\tabcolsep}{2pt}
\begin{tabular}{l|cccccc|c}
\toprule
\multirow{3}{*}{Method} & \multicolumn{2}{c}{MMD($\times10^2$)$\downarrow$} & \multicolumn{2}{c}{COV(\%)$\uparrow$} & \multicolumn{2}{c|}{1-NNA(\%)$\downarrow$} & \multirow{3}{*}{FID$\downarrow$} \\
\cmidrule(lr){2-3}\cmidrule(lr){4-5}\cmidrule(lr){6-7}
& CD & EMD & CD & EMD & CD & EMD \\
\midrule
DiffInDScene & 6.57 & 27.70 & 2.83 & 5.26 & 99.30 & 97.69 & 84.41 \\
BlockFusion & 2.90 & 28.79 & 16.60 & 13.16 & 97.89 & 98.19 & 25.09 \\
SynCity & 1.37 & 19.54 & 19.03 & 11.94 & 90.04 & 93.56 & 34.69 \\
TRELLIS & 3.15 & 23.75 & 13.97 & 11.74 & 99.20 & 98.79 & 53.49 \\
TRELLIS$^\dagger$ & 1.47 & 15.03 & 46.56 & 45.95 & 81.59 & 74.55 & 24.61 \\
\midrule
Ours & \textbf{0.97} & \textbf{13.33} & \textbf{51.82} & \textbf{46.56} & \textbf{66.30} & \textbf{69.01} & \textbf{7.52} \\
Ours w/o DC & 1.00 & 13.84 & 46.76 & 40.49 & 69.01 & 74.65 & 9.09 \\
Ours w/o CSG & 1.08 & 13.62 & 43.93 & 40.28 & 73.24 & 72.33 & 17.04 \\
\bottomrule
\end{tabular}
\vspace{-3pt}
\caption{Quantitative results on scene block geometry evaluation. We report comparisons with state-of-the-art scene generation methods, along with results from our ablation study. TRELLIS$^\dagger$ denotes TRELLIS fine-tuned on 3D-FRONT. ``DC'' refers to Data Curation, and ``CSG'' denotes Coarse Scene Generation.
}
\label{tab:single}
\end{table}

\paragraph{Implementation Details.}
We utilize a text-conditioned TRELLIS-XL model~\cite{trellis} for 3D block inpainting, where the conditioning text consists of a fixed generic scene description generated by a large language model~\cite{gpt4}. This prompt provides minimal semantic guidance, allowing the model to focus on spatial and structural reasoning.
For training, we optimize the inpainting model on our curated dataset for 200k iterations using AdamW~\cite{adamw} with a learning rate of 0.0001, while the VAE backbone is trained separately for 100k iterations with the same configuration. During inference, we maintain the same noise scheduling as training with 50 sampling steps. On a single A100 GPU, each block generation takes 20 seconds (6 times faster than SynCity's 2 minutes), and a complete 10$\times$10 indoor scene (around 272$\,\text{m}^2$) can be generated in 30 minutes using only 13GB of peak memory.

\begin{figure*}[thbp]
\centering
\includegraphics[width=\linewidth]{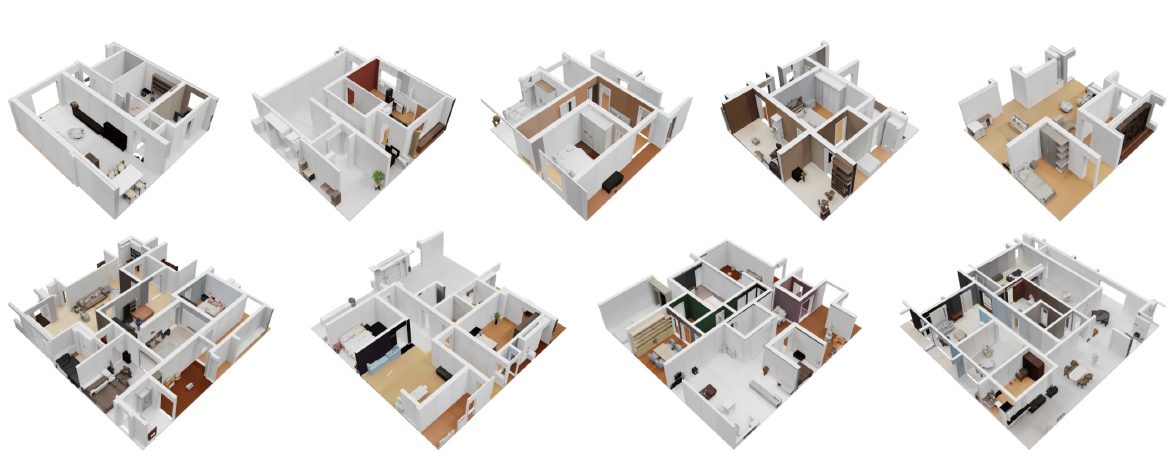}
\caption{Gallery of scenes generated by {WorldGrow}. Top: \(5\times5\) block layouts. Bottom: \(9\times9\) blocks.}
\label{fig:gallery}
\end{figure*}

\begin{figure*}[t]
\centering
\includegraphics[width=\linewidth]{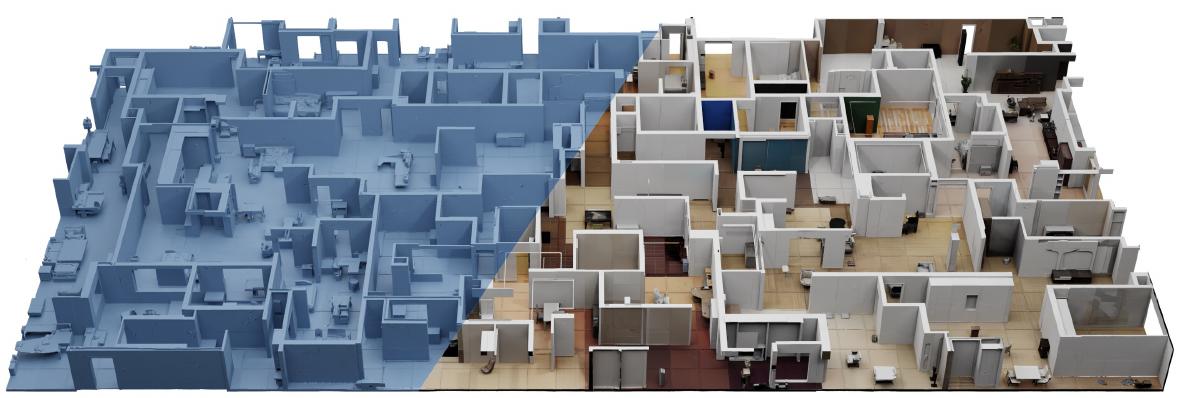}
\caption{Large-scale scene generated by WorldGrow. The layout spans \(19\times39\) blocks (\(\sim\)1{,}800\,m\(^2\)). Left: reconstructed mesh. Right: textured rendering. This example is distinct from Figure~\ref{fig:teaser} and illustrates scalability to large environments.}
\label{fig:gallery2}
\end{figure*}

\paragraph{Metrics.}
We evaluate our method across two aspects: scene block generation and full scene synthesis. 
For block generation, we assess both geometric and visual quality. Following prior works~\cite{blockfusion,lt3sd,diffusionsdf}, we report three standard distribution-based metrics (MMD, COV, and 1-NNA) computed using both Chamfer Distance (CD) and Earth Mover's Distance (EMD). We additionally adopt the perceptual Fréchet Inception Distance (FID)~\cite{fid} with PointNet++~\cite{pointnetpp} following the protocol in \cite{point-e} to assess 3D geometric quality. For visual quality, we render generated blocks from fixed multiple viewpoints and compute perceptual metrics including CLIP score~\cite{clip} and FID variants with different feature extractors (Inception V3~\cite{szegedy2016rethinking}, DINOv2~\cite{dinov2}, and CLIP).
For full-scene synthesis, where ground-truth meshes are unavailable, we conduct a human preference study with 91 participants who compare 5 methods across 10 scenes (4 house-level, 6 unbounded) presented in random order, evaluating structural plausibility, geometric detail, appearance fidelity, and scene continuity.

\begin{table}[t]
\centering
\small
\setlength{\tabcolsep}{5pt}
\begin{tabular}{l|cccc}
\toprule
Method & CLIP$\uparrow$ & FID$_{\text{Incep}}$$\downarrow$ & FID$_{\text{DINOv2}}$$\downarrow$ & FID$_{\text{CLIP}}$$\downarrow$ \\
\midrule
DiffInDScene$^\ddagger$ & 0.768 & 156.80 & 2066.13 & 42.43 \\
BlockFusion$^\ddagger$ & 0.758 & 138.34 & 1776.79 & 42.04 \\
SynCity & 0.804 & 101.83 & 655.60 & 16.22 \\
TRELLIS$^\dagger$ & 0.813 & 101.94 & 674.65 & 13.17 \\
\midrule
Ours & \textbf{0.843} & \textbf{29.87} & \textbf{313.54} & \textbf{3.95} \\
\bottomrule
\end{tabular}
\vspace{-3pt}
\caption{Visual fidelity evaluation of generated blocks. Methods with $^\ddagger$ generate geometry only; we apply uniform white texture for rendering and evaluation. TRELLIS$^\dagger$ denotes TRELLIS fine-tuned on 3D-FRONT.}
\label{tab:fidelity_eval}
\end{table}

\paragraph{Compared Methods.}
We compare \name with SOTA infinite scene generation methods, including BlockFusion~\cite{blockfusion} and SynCity~\cite{syncity}. Additionally, we evaluate against scene-scale generation baselines such as Text2Room~\cite{text2room} and DiffInDScene~\cite{diffindscene}. The original text-conditioned TRELLIS~\cite{trellis} is included as a foundational baseline for comparison.

\subsection{3D Scene Generation} \label{sec:compare}
\paragraph{Scene Block Generation.}
To evaluate scene block connectivity, we modify the evaluation protocol from BlockFusion and LT3SD. Instead of generating individual blocks in isolation, we task each method with synthesizing larger $3\times3$ scenes and randomly sample $1\times1$ blocks for evaluation against the 3D-FRONT dataset distribution.

As shown in Fig.~\ref{fig:qualitative}, SynCity exhibits poor continuity with visible discontinuities between segments, while other methods like fine-tuned TRELLIS produce locally valid blocks but lack outpainting capabilities. Quantitative results in Table~\ref{tab:single} confirm these observations, where \name achieves SOTA performance across all geometry metrics, demonstrating superior connectivity and structural coherence. 
To evaluate visual fidelity, we render the synthesized blocks from 10 fixed viewpoints and compare these multi-view images against renders from the 3D-FRONT dataset. As shown in Table~\ref{tab:fidelity_eval}, \name achieves significantly better perceptual quality than all baselines, demonstrating its ability to generate high-quality scene blocks with realistic appearance.

\paragraph{Full Scene Generation.}
We conduct a human preference study on textured indoor and unbounded scene generation. Following BlockFusion~\cite{blockfusion}, we ask participants to evaluate structure plausibility (SP), geometry detail (GD), and appearance fidelity (AF) for indoor scenes, with an additional criterion of continuity (CO) for unbounded scenes. As shown in Table~\ref{tab:userstudy}, our method outperforms baseline methods across all criteria, particularly excelling in scene structure layout and continuity—demonstrating the effectiveness of our block-by-block expansion and coarse-to-fine generation strategy. Fig.~\ref{fig:gallery} presents multiple distinct scenes produced by \name at increasing sizes, illustrating open-ended scalability across scene generation. We additionally present a \(19\times39\) indoor scene to highlight \name's scalability and consistency at large extents in Fig.~\ref{fig:gallery2}, demonstrating that \name can sustain quality when expanding far beyond the initial region, with minimal seams or drift, and yielding navigable, walk-only spaces suitable for planning-oriented embodied evaluation.

We also propose an expansion stability experiment to quantitatively assess long-run generation quality and error accumulation. In this experiment, we synthesize large scenes with 7$\times$7 blocks and randomly sample 1$\times$1 blocks exclusively from the outer regions (beyond the initial 3$\times$3 region) for evaluation using our block evaluation metrics. As shown in Table~\ref{tab:outer_eval}, \name maintains consistent generation quality even at distant expansions, achieving scores comparable to those in Table~\ref{tab:single}, while SynCity shows significant performance degradation (\eg, FID increases from 34.69 to 51.97). Note that SynCity fails in 70\% of expansion attempts, with only successful cases reported in the table. These results demonstrate \name's robust stability in infinite scene generation without quality deterioration or seam accumulation over extended expansions.

\begin{table}[t]
\centering
\small
\setlength{\tabcolsep}{5pt}
\begin{tabular}{l|ccc|cccc}
\toprule
\multirow{3}{*}{Method} & \multicolumn{3}{c|}{{Textured Scenes}} & \multicolumn{4}{c}{{Unbounded Scenes}} \\
\cmidrule(lr){2-4} \cmidrule(lr){5-8}
& SP & GD & AF & SP & GD & AF & CO \\
\midrule
Text2Room & 2.07 & 1.56 & 2.07 & / & / & / & / \\
Blockfusion & / & / & / & 3.48 & 3.30 & 1.20 & 3.36 \\
TRELLIS & 2.82 & 2.26 & 2.89 & 2.15 & 2.96 & 3.33 & 2.38 \\
SynCity & 2.48 & 3.11 & 3.59 & 2.48 & 3.07 & 4.08 & 2.74 \\
Ours & \textbf{4.48} & \textbf{4.44} & \textbf{4.33} & \textbf{4.46} & \textbf{4.37} & \textbf{4.33} & \textbf{4.69} \\
\bottomrule
\end{tabular}
\vspace{-3pt}
\caption{Average of human preference scores (1–5).}
\label{tab:userstudy}
\end{table}

\begin{table}[t]
\centering
\small
\setlength{\tabcolsep}{4pt}
\begin{tabular}{l|cccccc|c}
\toprule
\multirow{3}{*}{Method} & \multicolumn{2}{c}{MMD($\times10^2$)$\downarrow$} & \multicolumn{2}{c}{COV(\%)$\uparrow$} & \multicolumn{2}{c|}{1-NNA(\%)$\downarrow$} & \multirow{3}{*}{FID$\downarrow$} \\
\cmidrule(lr){2-3}\cmidrule(lr){4-5}\cmidrule(lr){6-7}
& CD & EMD & CD & EMD & CD & EMD \\
\midrule
SynCity & 1.68 & 19.39 & 15.38 & 13.97 & 94.27 & 93.76 & 51.97 \\
Ours & \textbf{0.96} & \textbf{12.83} & \textbf{48.99} & \textbf{48.18} & \textbf{59.66} & \textbf{64.79} & \textbf{5.43} \\
\bottomrule
\end{tabular}
\vspace{-3pt}
\caption{Expansion stability evaluation on outer regions. We evaluate 1$\times$1 blocks sampled from regions beyond the initial 3$\times$3 part among 7$\times$7 generated scenes. Our method maintains consistent quality in distant expansions, while SynCity shows significant degradation.}
\label{tab:outer_eval}
\end{table}

\subsection{Ablation Study} \label{sec:abl}
We perform a series of experiments to validate the effectiveness of each component. 

\begin{figure}[t]
\centering
\begin{subfigure}[b]{0.48\columnwidth}
    \centering
    \includegraphics[width=\textwidth]{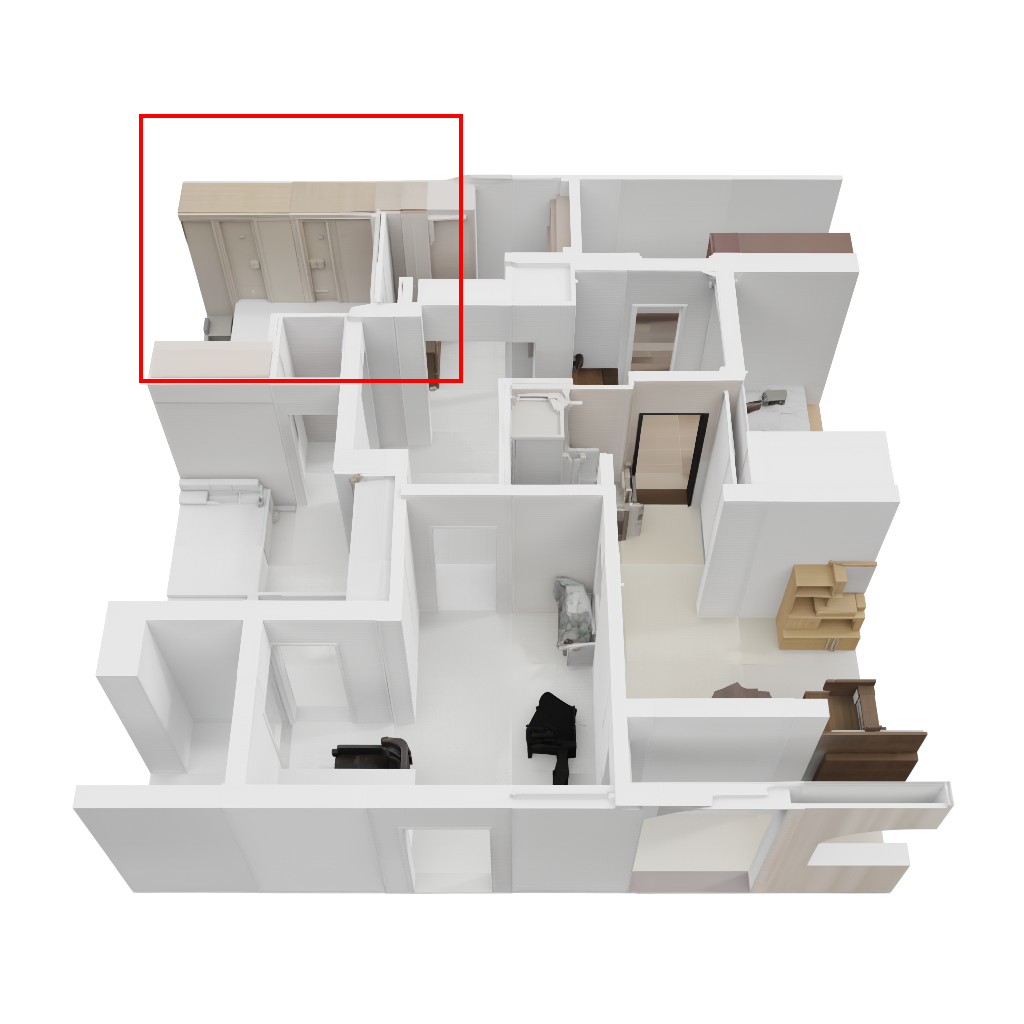}
    \label{fig:aba_data}
\end{subfigure}
\hfill
\begin{subfigure}[b]{0.48\columnwidth}
    \centering
    \includegraphics[width=\textwidth]{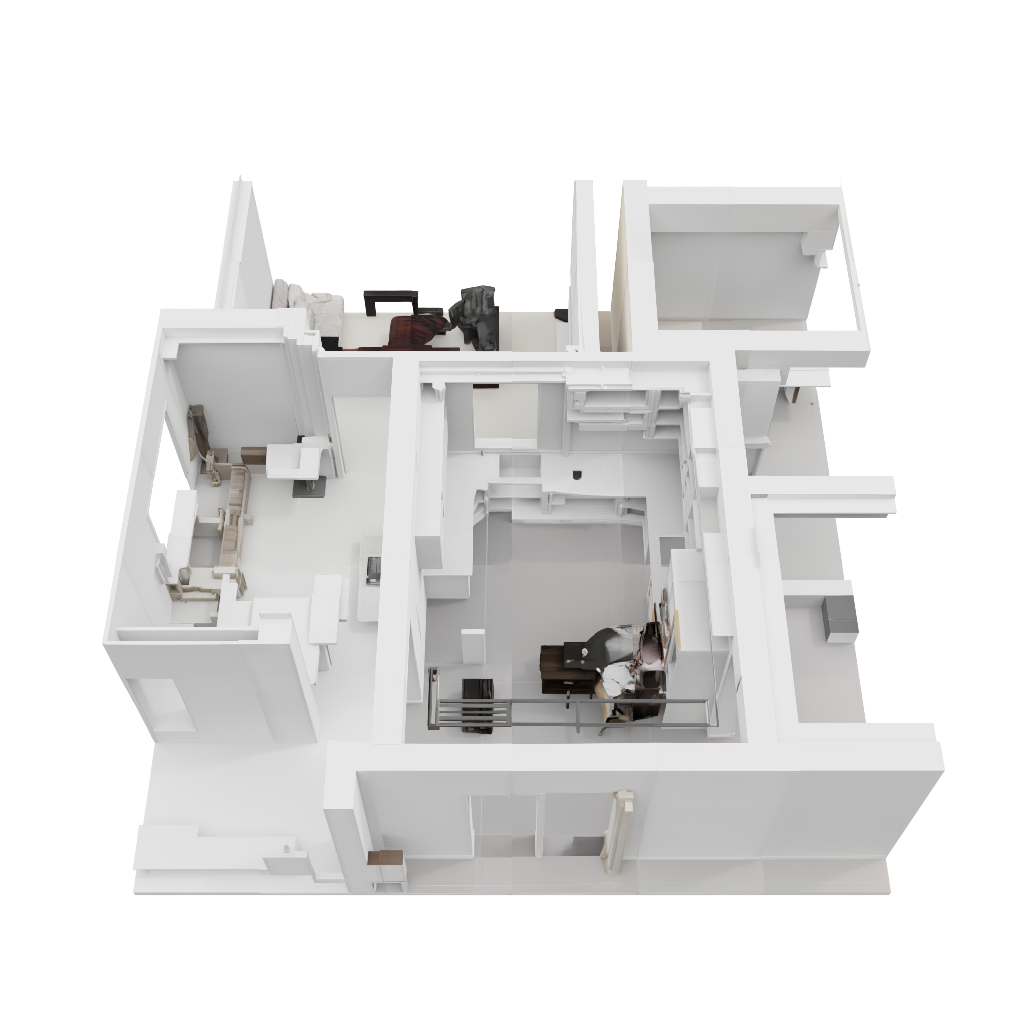}
    \label{fig:aba_c2f}
\end{subfigure}
\vspace{-2em}
\caption{Ablation study on key components of \name. Left: Without Data Curation, the generated wardrobe intersects with multiple walls, indicating poor spatial alignment. Right: Without Coarse-to-Fine generation, the global furniture layout becomes cluttered and less coherent.}
\label{fig:ablation}
\end{figure}

\paragraph{Data Curation.}
We first validate our data curation by comparing models trained on filtered versus unfiltered 3D-FRONT data. As shown in Fig.~\ref{fig:ablation}, training without data curation results in object interpenetration and implausible arrangements, while our curated dataset produces spatially coherent scenes.

\paragraph{Scene-Friendly SLAT.}
Our scene-friendly adaptation modifies TRELLIS's VAE to better support scene-level generation, introducing two key components: an occlusion-aware feature aggregation mechanism and a decoder retrained on scene blocks. To assess their impact on SLAT's ability to reconstruct realistic scene blocks, we conduct an ablation study against three variants: (i) the original object-centric VAE, (ii) a version with only occlusion-aware aggregation, and (iii) a version with only the retrained decoder.

As shown in Table~\ref{tab:vae}, applying occlusion-aware aggregation alone, without retraining the decoder, results in performance degradation due to encoder-decoder mismatch. However, combining two components yields significant improvements, demonstrating their synergy in adapting SLAT for coherent scene-level reconstruction.

\paragraph{Coarse-to-Fine Generation.}
Here, we validate our coarse-to-fine generation strategy by comparing against direct fine-scale generation. As shown in Fig.~\ref{fig:ablation}, direct fine generation struggles with global layout consistency, producing implausible furniture arrangements. Our coarse-to-fine approach establishes coherent structure via $\mathcal{G}_s^c$, then enriches details through $\mathcal{G}_s^f$, achieving superior balance between global coherence and local realism.

\begin{table}[ht]
\centering
\small
\setlength{\tabcolsep}{5pt}
\begin{tabular}{cc|ccc}
\toprule
Occ. Aware & Retrain $\mathcal{D}$ & {LPIPS} $\downarrow$ & {PSNR} $\uparrow$ & {SSIM} $\uparrow$ \\
\midrule
\ding{55} & \ding{55} & 0.0741 & 23.17 & 0.9273 \\
\ding{51} & \ding{55} & 0.0850 & 22.23 & 0.9046 \\
\ding{55} & \ding{51} & 0.0491 & 25.84 & 0.9531 \\
\ding{51} & \ding{51} & \textbf{0.0311} & \textbf{31.32} & \textbf{0.9705} \\
\bottomrule
\end{tabular}
\caption{Ablation study about components of scene-friendly SLAT. Occ. Aware means occlusion aware feature aggregation and Retrain $\mathcal{D}$ is retraining VAE's decoder.}
\label{tab:vae}
\end{table}

\section{Discussion and Future Work} \label{sec:lim}
While \name demonstrates strong results, several limitations remain. Currently, our method extends scenes only in the $XY$ plane, leaving vertical expansion along the $Z$-axis—essential for multi-story buildings—as an important direction for future work. Generation quality and diversity are also bounded by current 3D dataset limitations in scale, variety, and semantic annotations. Our block-wise design trades off fine geometric details for computational feasibility, prioritizing infinite generation capability over local detail resolution. Additionally, while \name naturally supports conditional control, the current implementation focuses on unconditional generation without semantic conditioning.

These points present clear opportunities for future research. Multi-level generation strategies could enable vertical expansion for complete buildings. Larger-scale dataset curation -- particularly for outdoor environments where our preliminary experiments on city scenes show promising results -- would enhance both diversity and quality. Introducing LLM-generated captions could enable fine-grained semantic control over room types and layouts. Moreover, it could be promising to integrate \name into geometry-appearance unified generation models~\cite{unilat3d} for more efficient pipelines.

\section{Conclusion} \label{sec:conclusion}
We presented \name, a novel framework for infinite 3D world generation that constructs unbounded environments with coherent layout and photorealistic appearance. Through our block-based context-aware inpainting mechanism and coarse-to-fine refinement strategy, we leverage pre-trained 3D priors to overcome the fundamental scalability and coherence limitations that have constrained prior methods. Our comprehensive evaluation demonstrates SOTA performance in geometry reconstruction and visual fidelity, while uniquely enabling the generation of large-scale scenes that maintain both local detail and global consistency. As virtual worlds become increasingly important for embodied AI training and simulation, \name provides a practical path toward scalable, high-quality 3D content generation for future world models.

{
    \small
    \bibliographystyle{ieeenat_fullname}
    \bibliography{main}
}

\clearpage
\appendix

\noindent In the appendix, we provide additional contents for:
\begin{enumerate}
    \item Method Details (Sec.~\ref{appdx:method})
    \item Implementation Details (Sec.~\ref{appdx:implement})
    \item Experiment Details (Sec.~\ref{appdx:exp})
    \item More Experimental Results (Sec.~\ref{appdx:results})
\end{enumerate}

\section{Method Details} \label{appdx:method}
\subsection{Coarse-to-Fine Generation}
\name introduces a hierarchical layout strategy where coarse and fine stages operate at different semantic levels rather than merely different resolutions. In our approach, each coarse block represents a $2\times2$ grid of fine blocks, capturing high-level spatial relationships and room-scale structure. The coarse model $\mathcal{G}_s^c$ establishes the global scene layout—defining where rooms connect and how spaces flow—while the fine model $\mathcal{G}_s^f$ fills in detailed geometry through structure-guided denoising. This hierarchical decomposition enables our model to reason about both global coherence (room-to-room relationships) and local detail (furniture, architectural elements) simultaneously. 
This design differs from prior coarse-to-fine approaches like LT3SD~\cite{lt3sd}, which uses a latent tree representation to encode both lower-frequency geometry and higher-frequency details in a multi-resolution hierarchy. While LT3SD models the same scene content at different resolutions, our method explicitly separates layout reasoning from detail generation, with each stage operating at a different semantic level. This hierarchical decomposition better aligns with the natural structure of indoor environments—where room arrangements constrain local details—enabling more coherent generation of large-scale scenes through explicit modeling of spatial relationships at multiple scales.

\subsection{Scene-Friendly SLAT}
Our occlusion-aware SLAT is specifically designed to handle the complex visibility patterns in indoor scenes, where walls, furniture, and architectural elements create extensive self-occlusion. Vanilla TRELLIS~\cite{trellis} was designed for single objects: it samples 150 views around an object and extracts pixel-wise DINOv2 features~\cite{dinov2} for each view. For every pixel, TRELLIS casts a ray and aggregates features along all voxels intersected by the ray, regardless of visibility. This approach works well for isolated objects but fails in scene contexts—features from a visible surface (\textit{e.g.}, a wall) get incorrectly projected onto occluded voxels behind it (\textit{e.g.}, furniture in another room), causing severe artifacts and inconsistent appearance generation.

In \name, we introduce an occlusion-aware visual feature aggregation strategy to address this issue. To compute the sparse voxel features $\mathbf{f} = \{(\mathbf{f}_i, \mathbf{p}_i)\}_{i=1}^L$, each voxel center $\mathbf{p}_i$ is projected onto multiple camera views, where we compute binary visibility masks $\{M_v\}$ using depth testing. The occlusion-aware feature $\mathbf{f}_i$ is then computed by averaging DINOv2 features from only the views where $\mathbf{p}_i$ is visible according to $M_v$. This ensures that each voxel only receives features from views where it is actually observable, preventing feature contamination across occluded surfaces.

This occlusion-aware lifting significantly improves representation quality when integrated into the TRELLIS encoder $\mathcal{E}$. Consider walls with distinct textures on opposite sides—the original method would incorrectly blend features from both sides, while our approach preserves these distinctions, resulting in sharper boundaries and more accurate appearance modeling.

To fully leverage these improved features, we retrain the SLAT decoder $\mathcal{D}$ on scene-scale data with occlusion-aware features $\mathbf{f}$. This retraining adapts the decoder to the unique challenges of scene generation, improving photorealism, preserving geometric boundaries, and ensuring coherent appearance across block transitions—critical capabilities for high-quality unbounded scene synthesis.

\section{Implementation Details} \label{appdx:implement}
\subsection{Datasets}
We curate a high-quality subset from 3D-FRONT~\cite{3dfront} by filtering out houses with mesh penetration (1,971), incorrect furniture placement (1,232), small layouts (324), sparse furnishing (456), and other anomalies (585), yielding 3,072 clean houses. We additionally select 353 houses with minor issues for manual refinement. From these datasets, we extract 100K fine and 30K coarse blocks from the filtered houses, plus 20K fine and 8K coarse blocks from the manually refined subset. Our models are first trained on the larger filtered set, then fine-tuned on the curated subset for improved quality.

For block extraction, we use a standard house height of $h \approx 3\text{m}$. Fine blocks are cubes with width $w^f = h$, while coarse blocks span $w^c = 2h$ (covering a $2\times2$ grid of fine blocks), both maintaining height $h$.

To evaluate generalization to outdoor scenes, we train \name on UrbanScene3D~\cite{urbanscene3d}. Among its seven cities, we select \textit{Shanghai} for its architectural diversity and texture quality. We extract 10K fine and 3K coarse blocks, with each fine block covering $100\text{m}$—suitable for long-range outdoor synthesis.

\subsection{Text Prompts}
We use the following prompt for indoor scene generation:
\begin{quote}
\textsl{
A photorealistic 3D house mesh with contemporary architectural style, featuring clean lines and a balanced mix of natural and industrial materials. Use large windows for natural lighting. Prioritize smooth mesh topology and modular components for adaptability, maintaining a cohesive modern aesthetic.}
\end{quote}
For outdoor scenes:
\begin{quote}
\textsl{
A realistic 3D urban street scene in daylight, featuring modern buildings, parked cars, street lamps and sidewalks. The environment should be detailed and clean.}
\end{quote}
Both prompts were generated using GPT-4~\cite{gpt4} and encoded with a frozen CLIP~\cite{clip} text encoder, following TRELLIS-text-xlarge's pipeline.

\section{Experiment Details} \label{appdx:exp}
\subsection{Metrics}
We implement MMD, COV, and 1-NNA following NFD~\cite{nfd}. For FID~\cite{fid}, we diverge from LT3SD's and TRELLIS's image-based evaluation, which renders views and computes FID on 2D images. 
We also compute FID on 3D mesh data using PointNet++~\cite{pointnetpp} features, ensuring fair comparison across all methods.

\subsection{Compared Methods}
We compare \name with representative scene-scale generators spanning image lifting, block-wise extrapolation, diffusion-based volumetric synthesis, and TRELLIS-family baselines. Text2Room~\cite{text2room} reconstructs 3D scenes by lifting multi-view 2D images via depth estimation and iterative fusion. BlockFusion~\cite{blockfusion} grows scenes block-by-block through latent tri-plane extrapolation. DiffInDScene~\cite{diffindscene} employs a cascaded diffusion pipeline to produce room-level TSDFs that are fused locally. Vanilla TRELLIS~\cite{trellis} serves as our base model; we also report a fine-tuned variant trained on our curated 3D-FRONT subset, denoted TRELLIS$^\dagger$. SynCity~\cite{syncity} is a training-free TRELLIS variant that couples TRELLIS geometry with 2D generators for large-scale synthesis. We exclude LT3SD~\cite{lt3sd} due to the absence of released checkpoints and training details necessary for reproducible evaluation. We also omit WonderWorld, which optimizes from a small set of given 2D views and does not produce a globally consistent 3D scene.

\begin{table}[thbp]
\centering
\small
\setlength{\tabcolsep}{2pt}
\begin{tabular}{l|cccccc|c}
\toprule
\multirow{3}{*}{Method} & \multicolumn{2}{c}{MMD($\times10^2$)$\downarrow$} & \multicolumn{2}{c}{COV(\%)$\uparrow$} & \multicolumn{2}{c|}{1\text{-}NNA(\%)$\downarrow$} & \multirow{3}{*}{FID$\downarrow$} \\
\cmidrule(lr){2-3}\cmidrule(lr){4-5}\cmidrule(lr){6-7}
& CD & EMD & CD & EMD & CD & EMD \\
\midrule
SynCity & 0.42 & 6.78 & 29.00 & 34.80 & 95.30 & 90.00 & 93.45 \\
Ours & \textbf{0.41} & \textbf{6.35} & \textbf{41.80} & \textbf{44.80} & \textbf{81.30} & \textbf{84.40} & \textbf{23.49} \\
\bottomrule
\end{tabular}
\caption{Outdoor/urban scene generation on \textit{UrbanScene3D}~\cite{urbanscene3d}. We train \name on the \textit{Shanghai} split using 10K fine and 3K coarse blocks (each fine block covers $100\,\mathrm{m}$). Even with this small training set, \name achieves markedly better coverage and perceptual quality than SynCity.}
\label{tab:outdoor}
\end{table}

\section{Additional Results} \label{appdx:results}

\subsection{Outdoor 3D Scene Generation}
To assess outdoor performance, we train \name on \textit{UrbanScene3D}~\cite{urbanscene3d}. Among its seven cities, we select \textit{Shanghai} for its architectural diversity and texture quality. As shown in Fig.~\ref{fig:outdoor} and Table~\ref{tab:outdoor}, \name attains comparable or better geometric statistics than SynCity (lower MMD and 1\text{-}NNA; higher COV) and substantially improves perceptual quality (FID: 23.49 vs.\ 93.45), indicating that our coarse-to-fine refinement and masked conditioning remain effective in large-scale urban layouts.

Given the limited size of the training subset, these results should be viewed as preliminary but encouraging rather than a conclusive benchmark. We expect further gains from larger and more diverse outdoor datasets, \textit{e.g.}, city-scale aerial (drone or satellite) and street-level captures—as well as highly diverse synthetic assets from procedural systems such as Infinigen~\cite{raistrick2023infinite}.

\end{document}